\documentclass[runningheads]{llncs}
\usepackage{subcaption} 

\usepackage{amsmath}     
\usepackage{amssymb} 
\usepackage{graphicx}
\usepackage{xcolor}

\usepackage{rotating}

\begin{document}
\title{Evaluating Prompting Strategies for \\ Chart Question Answering with \\ Large Language Models}
\titlerunning{Evaluating Prompting Strategies for Chart QA}
%
\author{Ruthuparna Naikar \and
Ying Zhu}
\authorrunning{Naikar and Zhu}
%
\institute{Georgia State University, Atlanta GA, USA\\ \email{ruthuparna4@gmail.com, yzhu@gsu.edu}
}
\maketitle              
\begin{abstract}

Prompting strategies affect LLM reasoning performance, but their role in chart-based QA remains underexplored. We present a systematic evaluation of four widely used prompting paradigms (Zero-Shot, Few-Shot, Zero-Shot Chain-of-Thought, and Few-Shot Chain-of-Thought) across GPT-3.5, GPT-4, and GPT-4o on the ChartQA dataset. Our framework operates exclusively on structured chart data, isolating prompt structure as the only experimental variable, and evaluates performance using two metrics: Accuracy and Exact Match. Results from 1,200 diverse ChartQA samples show that Few-Shot Chain-of-Thought prompting consistently yields the highest accuracy (up to 78.2\%), particularly on reasoning-intensive questions, while Few-Shot prompting improves format adherence. Zero-Shot performs well only with high-capacity models on simpler tasks. These findings provide actionable guidance for selecting prompting strategies in structured data reasoning tasks, with implications for both efficiency and accuracy in real-world applications.
\end{abstract}


\section{Introduction}

Large Language Models (LLMs) have demonstrated strong capabilities in natural language understanding, reasoning, and generation across domains. Recently, they have been applied to tasks beyond textual inputs, including data visualization, semantic interpretation, and  Question Answering (QA) over structured information. 
A challenging variant is \textit{chart-based} question answering (Chart QA), where the model must answer questions by reasoning over structured data extracted from data visualizations such as bar charts, line graphs, or scatter plots.

While multimodal approaches integrate visual parsing with symbolic reasoning \cite{khan2025evaluating,huang2025vprochart,lee2023pix2struct}, an emerging line of work treats Chart QA as a structured text reasoning problem by converting charts into tabular or serialized textual representations. This approach bypasses the complexities of optical character recognition (OCR) and visual feature extraction, instead leveraging the inherent reasoning abilities of LLMs over structured text.

A critical yet underexplored factor in this paradigm is the design of \textit{prompts} used to guide LLMs. Prompting strategies vary in complexity, ranging from minimal Zero-Shot instructions to demonstration-based Few-Shot prompting to reasoning-augmented variants such as Zero-Shot Chain-of-Thought (ZS-CoT) and Few-Shot Chain-of-Thought (FS-CoT). These strategies can substantially influence model accuracy, consistency, and cost trade-offs. 
While prompt engineering has been extensively studied in textual tasks, its systematic evaluation for structured chart reasoning remains limited.

In this work, we address this gap by:
\begin{itemize}
    \item Isolating the effect of four prompting strategies (Zero-Shot, Few-Shot, Zero-Shot Chain-of-Thought, and Few-Shot Chain-of-Thought) on Chart QA performance, treating the LLM as a fixed black box.
    \item Conducting experiments on the ChartQA benchmark using exclusively structured tabular inputs, ensuring controlled comparisons without visual input confounds.
    \item Evaluating two metrics, Accuracy and Exact Match, on three LLM variants with different capability and efficiency profiles: GPT-3.5, GPT-4, and GPT-4o.
\end{itemize}

Our results provide empirical guidance for prompt selection in structured data reasoning tasks, showing that FS-CoT delivers the highest reasoning accuracy, Few-Shot improves adherence to expected formats, ZS-CoT offers a middle ground by enhancing reasoning without exemplars, and Zero-Shot remains viable for high-capacity models on simpler queries. 
This work contributes to both the methodological understanding of prompt engineering and its practical application in chart-based visual reasoning.

\section{Related Work}

\subsection{LLMs in Data Visualizations}

Large Language Models (LLMs) have been applied to diverse visualization tasks, including chart generation, semantic interpretation, and question answering (QA) \cite{wu2024automated,LiChen2025,chen2025viseval}. For example, ChartGPT~\cite{10443572} can generate full chart specifications directly from natural language.  

Other efforts, such as Khan et al. \cite{khan2025evaluating} and \textit{VProChart}~\cite{huang2025vprochart}, show the potential of LLMs for chart interpretation but still rely on complex multimodal pipelines or code synthesis. More general multimodal models have also pushed this direction forward. Pix2Struct \cite{lee2023pix2struct}, for example, converts chart images into structured text that LLMs can reason over. More recently, Wu et al.~\cite{wu2024chartinsights} propose \textit{ChartInsights}, a benchmark for low-level chart QA, showing that even advanced multimodal models underperform on basic analytical tasks.

Together, these works demonstrate the versatility of LLMs in visualization but concentrate primarily on generation, captioning, or multimodal interpretation. In contrast, our work focuses on structured chart QA, treating LLMs as black-box inference engines and systematically isolating the effect of prompt design on reasoning performance.

\subsection{Prompt Strategies for Large Language Models}

Prompt engineering has emerged as a practical alternative to fine-tuning for adapting LLMs to downstream tasks. Debnath et al.~\cite{Debnath_2025} present a survey of over fifty techniques, spanning instructional prompts, zero-shot exemplars, chain-of-thought reasoning, and automatic optimization. Liu et al.~\cite{liu20comprehensive} present a comprehensive taxonomy of prompt engineering techniques for large language models, categorizing methods into manual, automatic, and optimization-based strategies. While these surveys cover tasks such as classification, summarization, and multi-hop reasoning, their scope remains largely textual, with limited attention to structured domains like chart QA.

Zhu et al.~\cite{zhu2024promptbench} introduce \textit{PromptBench}, a unified benchmark for comparing prompt engineering strategies across tasks and model families.  
Wei et al.~\cite{NEURIPS2022_9d560961} propose Chain-of-Thought (CoT) prompting, demonstrating that step-by-step reasoning improves performance on complex tasks.  
Wang et al.~\cite{wang2022self} extend this with \textit{Self-Consistency}, showing that sampling diverse reasoning paths and selecting the most frequent answer boosts reliability.

However, none of these studies systematically examines how prompting strategies influence reasoning over structured data. Our work addresses this gap by evaluating general-purpose LLMs under four paradigms (Zero-Shot, Few-Shot, Zero-Shot CoT, and Few-Shot CoT), keeping models constant and varying only the prompt template. This design directly isolates how prompting strategies shape accuracy and consistency in chart QA.

\subsection{Visual Question Answering Over Charts}

Chart-based Question Answering (Chart QA) is a specialized branch of QA that requires reasoning over structured visualizations such as bar charts, line graphs, and infographics. Unlike object-centric QA, Chart QA demands precise interpretation of axes, legends, and encodings, often involving arithmetic, aggregation, or comparative reasoning.  

Kafle et al.~\cite{kafle2018dvqa} introduce \textit{DVQA}, featuring synthetic bar charts with questions targeting symbolic reasoning. Methani et al.~\cite{methani2020plotqa} extend this with \textit{PlotQA}, a larger dataset of scientific plots, though both rely heavily on OCR and handcrafted components, making them sensitive to layout variation and textual noise. Masry et al.~\cite{masry2022chartqa} propose \textit{ChartQA}, combining real and synthetic charts to support multi-step reasoning, though its transformer-based architecture remains resource-intensive. \textit{InfographicVQA}~\cite{mathew2022infographicvqa} broadens the scope to dense infographics mixing complex layouts with embedded text.  

More recent approaches shift from pixel-level processing to abstraction. \textit{DePlot}~\cite{liu2022deplot} converts chart images into table-formatted text, enabling LLMs to reason over visual content via textual representations. Similarly, Liu et al.~\cite{liu2024chartqa} serialize charts into structured sequences for LLM processing, showing that careful formatting and prompting can outperform multimodal baselines. Deng et al.~\cite{deng2024tables} compare text- and image-based representations for table reasoning, finding that representation choice substantially impacts LLM performance.  

These works establish chart QA as a promising research area but do not isolate the role of prompt design. Our work builds on this abstraction trend by operating directly on structured chart representations with instruction-tuned LLMs and systematically comparing prompting strategies (Zero-Shot, Zero-Shot CoT, Few-Shot, and Few-Shot CoT), highlighting their distinct effects on reasoning depth, output consistency, and generalization without reliance on visual parsing or fine-tuning.  

\section{Methodology}

In this section, we describe the architecture and design principles underlying our system for answering questions over structured chart data using large language models (LLMs).

\subsection{Large Language Models and Prompting}

Large Language Models (LLMs) such as GPT-3.5, GPT-4, and GPT-4o~\cite{openai2023gpt4} are transformer-based neural networks trained on large-scale text corpora with strong generalization to structured reasoning tasks. In this study, we use them in an inference-only setting via the OpenAI API. The model behavior is controlled through natural language prompts, which embed task instructions, input data, and queries into a single text string (e.g., a serialized chart paired with a question). Rather than training custom models or designing multimodal pipelines, we treat LLMs as black-box reasoning engines and isolate the role of prompt design in shaping model behavior. 

\subsection{Architecture Overview}

Our framework is modular and prompt-centric, operating on structured textual chart representations rather than raw images. Instruction-tuned LLMs are used as black-box inference engines, with an architecture consisting of four components: (1) Input Representation, (2) Prompt Generation, (3) LLM Inference, and (4) Answer Extraction and Evaluation. This design enables controlled comparison of prompting strategies while holding model configurations and inputs constant.

\begin{figure}[ht]
    \centering
    \includegraphics[width=0.95\linewidth]{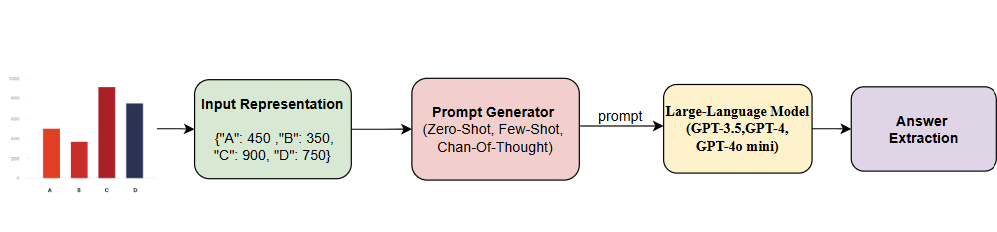}
    \caption{This diagram illustrates the workflow of chart-based question answering with LLMs. Structured chart data and natural language questions are transformed into prompts, which are submitted to an instruction-tuned LLM to generate answers.}
    \label{fig:system_pipeline}
\end{figure}

\begin{itemize}
    \item \textbf{Input Representation} — Each instance includes a chart serialized into a tabular key–value or CSV-like format and a natural language question, sourced from the \textit{ChartQA} dataset \cite{masry2022chartqa}.
    
    \item \textbf{Prompt Generation} — A formatting module combines the serialized chart data and question into a complete prompt following one of three strategies: Zero-Shot, Few-Shot, or Chain-of-Thought (CoT).
    
    \item \textbf{LLM Inference} — The prompt is submitted via API to an instruction-tuned LLM (GPT-3.5, GPT-4, or GPT-4o) in inference-only mode. Outputs are returned in natural language without fine-tuning or architectural changes.
    
    \item \textbf{Answer Extraction and Evaluation} — The generated response is parsed to extract a candidate answer and compared to the ground truth using exact match and other metrics. Additional diagnostics assess consistency, reasoning depth, and error types.
\end{itemize}

This architecture is model-agnostic, lightweight, and interpretable, enabling transparent analysis of LLM behavior under varying prompt configurations. All intermediate artifacts—including prompts, structured inputs, and model outputs—are human-readable, supporting reproducibility and qualitative inspection.

\subsection{Data} \label{data}

We conducted experiments using the ChartQA dataset~\cite{masry2022chartqa}, a widely used benchmark for evaluating the answer to questions on charts. Designed to test symbolic and quantitative reasoning, ChartQA contains approximately 28{,}000 question-answer pairs linked to 7{,}000 unique charts, evenly divided between synthetic and real-world sources. Each chart is associated with 1--5 natural language questions of varying linguistic form and reasoning complexity. For our experiments, we sampled 1,200 question–answer pairs, comprising 600 human-authored and 600 augmented (synthetically generated) questions, to cover a broad spectrum of reasoning types. 

\begin{figure}[h]
    \centering
    \begin{subfigure}{0.48\textwidth}
        \centering
        \includegraphics[width=\linewidth]{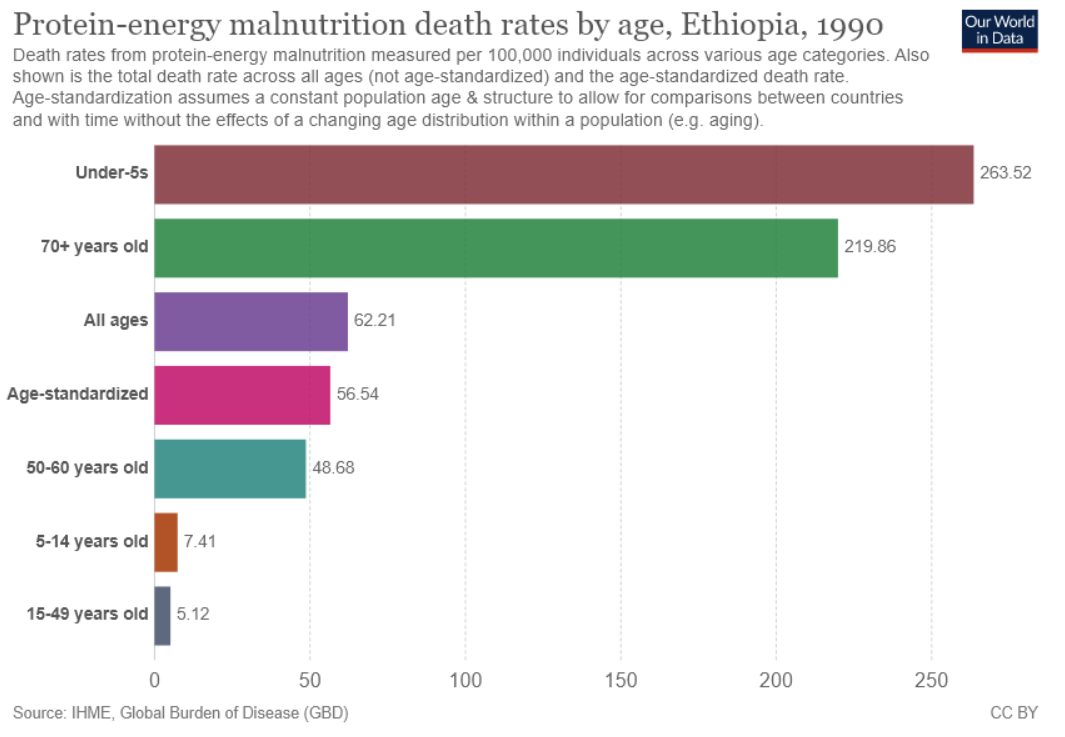}
        \caption{Visualization of the sample data}
        \label{fig:graph}
    \end{subfigure}\hfill
    \begin{subfigure}{0.48\textwidth}
        \centering
        \includegraphics[width=\linewidth]{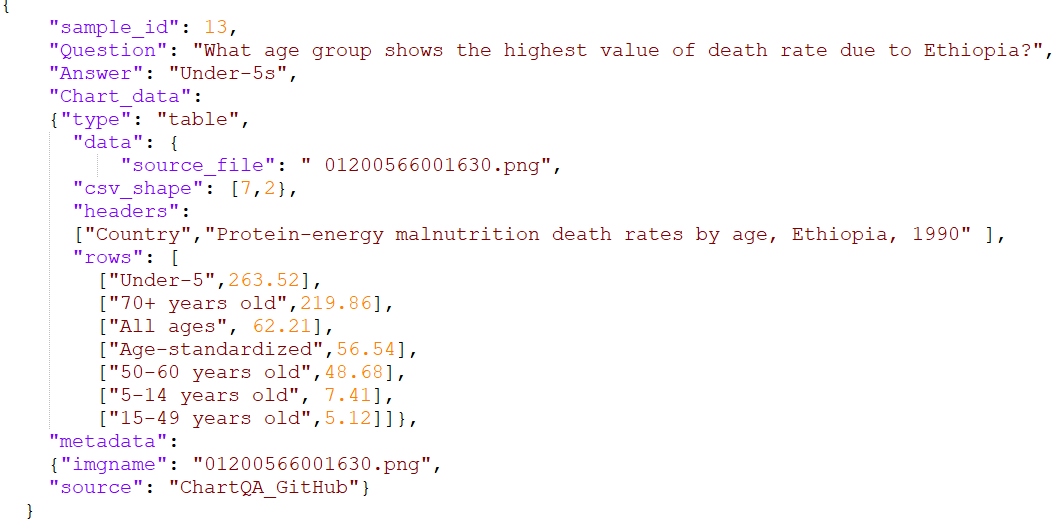}
        \caption{Serialized JSON from the same table data with question–answer pair}
        \label{fig:json}
    \end{subfigure}
    \caption{Sample data visualization and its serialized JSON representation.}
    \label{fig:graph_json}
\end{figure}

The questions span multiple reasoning types:
\begin{itemize}
    \item \textbf{Arithmetic}: \textit{``What is the difference between the death rates of ``Under-5” and ``70+ years old?"}
    \item \textbf{Comparative reasoning}: \textit{``Which age group has the lowest death rate? "}
    \item \textbf{Boolean}: \textit{``Is the death rate for ``Under-5” higher than 200?"}
    \item \textbf{Direct Retrieval}:\textit{``What is the death rate for ``Age-standardized"?"}
\end{itemize}

Since our study focuses on structured textual reasoning rather than visual parsing, we focus solely on tabular data and QA pairs. We exclude questions that rely on purely visual cues (e.g., colors, shapes, counts of graphical elements), as these require the rendered chart and are outside our scope. Preprocessing links each question to its corresponding table using the shared image filename, converts the CSV into a row-wise key–value format, and combines it with the question to form a complete model prompt. This representation preserves the structural integrity of the data, allowing controlled evaluation of prompt design across diverse chart types and reasoning tasks. 

\subsection{Model Selection} \label{model_selection}

We evaluated three OpenAI LLMs~\cite{openai2023gpt4} to compare prompting strategies across different capability levels, while maintaining consistency within the same model family. This design controls for confounding variables, such as vendor differences or pretraining corpora, allowing observed performance variations to be primarily attributed to prompt design. 

\begin{itemize}
    \item \textbf{GPT-3.5} — A widely used baseline with lower cost and weaker reasoning, included to examine how prompting strategies compensate for limited capability. 
    \item \textbf{GPT-4} — The most capable variant, optimized for advanced reasoning and factual accuracy. 
    \item \textbf{GPT-4o} — A smaller, faster, and more cost-efficient GPT-4 variant, used to assess the impact of prompting on efficiency-oriented models. 
\end{itemize}

\subsection{Experimental Setup} \label{experimental_setup}

Using the dataset described in Section~\ref{data}, we evaluate four prompting strategies: Zero-Shot (ZSP), Zero-Shot Chain-of-Thought (ZS-CoT), Few-Shot (FSP), and Few-Shot Chain-of-Thought (FS-CoT) — on GPT-3.5, GPT-4, and GPT-4o via the OpenAI Chat Completions API. All runs use identical decoding settings to ensure determinism and comparability: temperature = 0, top-p = 1, a 1{,}024-token output limit, and fixed stop sequences. The preprocessing and evaluation splits are kept constant across the strategies.

For analysis, we group questions into four reasoning categories—Arithmetic, Comparative, Boolean, and Direct Retrieval—assigned using a deterministic rule-based classifier. When multiple rules apply, we use the precedence order: Arithmetic $\rightarrow$ Comparative $\rightarrow$ Boolean $\rightarrow$ Direct Retrieval. 
\begin{itemize}
  \item \textbf{Arithmetic}: requires explicit computation (e.g., sum, difference, average, ratio, change).
  \item \textbf{Comparative}: requires ordering or argmax/argmin (e.g., higher, lower, most, least, rank).
  \item \textbf{Boolean}: yes/no or threshold-based without multi-step arithmetic (e.g., “is,” “at least,” “more than”).
  \item \textbf{Direct Retrieval}: lookup of one or a few table cells given keys (entity, year, attribute).
\end{itemize}
A manual check of 100 random samples yielded 99\% classification accuracy. 
To construct a balanced evaluation set of 1{,}200 items (300 per category), we classified over 6{,}000 randomly sampled ChartQA questions and retained the first 300 per category. Boolean and Arithmetic required oversampling due to scarcity.

For Few-Shot and FS-CoT prompting, we maintain a fixed pool of 50 exemplars spanning the four categories (12–13 per category). 
In inference, three exemplars from the same category as the target question are uniformly sampled without replacement (with a fixed seed for reproducibility) and prepended to the prompt. 
In FS-CoT, exemplars include concise step-by-step rationales, while the target includes only the reasoning scaffold. 
To prevent leakage, exemplars are drawn from non-overlapping charts outside the evaluation split.

API usage is uniform across models: authentication via \texttt{OPENAI\_API\_KEY}, one completion per prompt, no tool/function calling or retrieval, and seed = 0 where supported. We pre-generate and freeze all 1{,}200 prompts (table + question + context when applicable) and execute them in batches across models, ensuring strict comparability.

\subsection{Prompting Strategies} \label{prompt_stratergies}
Prompting enables LLMs to adapt to new tasks in inference-only settings where parameters remain fixed. 
We evaluate four primary strategies:
\begin{itemize}
  \item \textbf{Zero-Shot (ZSP)}: minimal task instruction without examples or reasoning triggers.
  \item \textbf{Few-Shot (FSP)}: prepends three same-category exemplars without reasoning traces. 
  \item   \textbf{Zero-Shot CoT (ZS-CoT)}: adds a reasoning scaffold (e.g., “Let’s think step by step”) without exemplars. 
  \item \textbf{Few-Shot CoT (FS-CoT)}: prepends exemplars with concise step-by-step rationales plus a reasoning scaffold in the target.
\end{itemize}
This design isolates the effect of prompt structure on model accuracy and consistency in structured chart reasoning.

 \subsubsection{Prompt Construction}
Each item consists of a chart-derived table and a natural-language question. We serialize tables to plain text (row-wise, CSV-like) with normalized column names and units so they are unambiguous for text-only inference. A single instruction preamble is used throughout: use only the provided table; return the final answer after \texttt{Answer:}; numeric outputs as bare numbers (round to 2 decimals if needed); booleans as \texttt{Yes}/\texttt{No}; categorical answers as the exact cell string.
\begin{figure}[ht]
    \centering
    \includegraphics[width=0.8\linewidth]{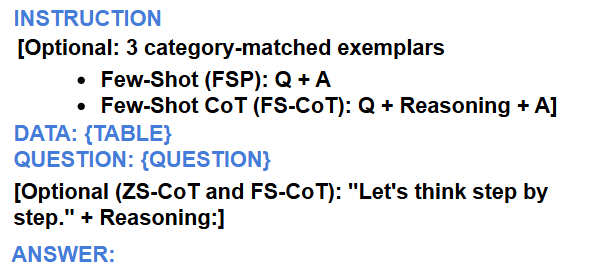}
    \vspace{-.5em} 
    \caption{Generalized prompt template with components.}
    \label{fig:prompt_template}
\end{figure}

Prompts are constructed once for all 1{,}200 items and reused verbatim across models. Unlike textual QA, chart reasoning introduces diverse reasoning types: arithmetic, comparative, Boolean, and direct retrieval, which combine symbolic computation with natural language understanding. Prompting strategies effective in purely textual tasks may therefore behave differently in this domain. By isolating prompt design under structured chart representations, our study explicitly targets this domain-specific gap in prior prompting benchmarks.

\subsubsection{Prompt Variants}
\begin{itemize}
  \item \textbf{Zero-Shot} Presents only the serialized table and the target question; the prompt ends with \texttt{ Answer:}.
  \item \textbf{Zero-Shot CoT (ZS-CoT)} Same as Zero-Shot, but appends a single CoT trigger (e.g., ``Let's think step by step.'') and a blank \texttt{Reasoning:} line before \texttt{Answer:}; no exemplars are provided.
  \item \textbf{Few-Shot} Prepends three \emph{category-matched} exemplars sampled uniformly (without replacement; fixed RNG seed) from an auxiliary pool of 50 items. The auxiliary pool is disjoint from the evaluation split (no shared images, tables, or questions).
  \item \textbf{Few-Shot Chain-of-Thought (FS-CoT)} Identical to Few-Shot, but each exemplar includes a concise, standardized reasoning trace; the target includes a blank \texttt{Reasoning:} scaffold before \texttt{Answer:}.
\end{itemize}
This design keeps the serialized-table core constant and varies only the presence and form of contextual guidance, isolating the effect of prompt design.

\subsubsection{Prompt Examples}
To illustrate the prompting strategies, we show real examples from our evaluation datasets.

\begin{table}[ht]
\centering
\caption{Example of Few-Shot prompting with structured chart data and model output.}
\label{tab:fewshot_example}
\resizebox{\textwidth}{!}{%
\begin{tabular}{|p{3cm}|p{9cm}|}
\hline
\textbf{Component} & \textbf{Content} \\ \hline
\textbf{Instruction} &
Use the provided table to answer the question. Respond only with the final answer after ``\texttt{Answer:}''. Round numbers to two decimals. \\ \hline
\textbf{Exemplar 1} &
\textbf{Table:} Year–GDP (trillion USD): 2018–2.1, 2019–2.3.  
\textbf{Q:} What is the GDP growth from 2018 to 2019?  
\textbf{Answer:} 0.2 \\ \hline
\textbf{Exemplar 2} &
\textbf{Table:} Country–Literacy Rate (\%): India–74, China–96.  
\textbf{Q:} Which country has a higher literacy rate?  
\textbf{Answer:} China \\ \hline
\textbf{Exemplar 3} &
\textbf{Table:} Year–Death Rate (Under-5): 2000–100, 2010–60.  
\textbf{Q:} What is the difference in death rate between 2000 and 2010?  
\textbf{Answer:} 40 \\ \hline
\textbf{Target Question (Data + Question)} &
\textbf{Table:} Age Group–Death Rate: Under-5–170.2, 15–49–95.7, 70+–610.5.  
\textbf{Q:} Which age group has the lowest death rate?  
\textbf{Model Output (GPT-4o):} 15–49  
\textbf{Expected Answer:} 15–49 \\ \hline
\end{tabular}%
}
\end{table}

\begin{table}[ht]
\centering
\caption{Example of Zero-Shot prompting with structured chart data and model output.}
\label{tab:zeroshot_example}
\resizebox{\textwidth}{!}{%
\begin{tabular}{|p{3cm}|p{9cm}|}
\hline
\textbf{Component} & \textbf{Content} \\ \hline

\textbf{Instruction} &
Use the provided table to answer the question.  
Respond only with the final answer after ``\texttt{Answer:}''.  
Round numbers to two decimals. \\ \hline

\textbf{Target Question (Data + Question)} &
\textbf{Table:} Age Group–Death Rate:  
Under-5–170.2, 15–49–95.7, 70+–610.5.  
\textbf{Q:} Which age group has the lowest death rate?  
\textbf{Model Output (GPT-4o):} 15–49  
\textbf{Expected Answer:} 15–49 \\ \hline

\end{tabular}%
}
\end{table}

\subsubsection{Prompt Output Handling and Post-processing}
Model outputs were stored together with their corresponding prompts to maintain complete traceability. Minor formatting variations (such as extra spaces, inconsistent punctuation, or line breaks) were normalized to align with the gold labels. The resulting standardized outputs were compiled into an evaluation matrix, providing a consistent basis for all quantitative and qualitative analyses.

\section{Results}\label{evaluation}

Once all the OpenAI responses were generated, we conducted an extensive evaluation across the three GPT models and the four prompting strategies. The objective was to measure how each combination performed in Chart QA tasks and to identify trends in model–prompt interaction. The evaluation was conducted on a dataset of 1,200 Chart QA samples. Two evaluation metrics were used: Accuracy and Exact Match (EM). These metrics together provide a balanced assessment, covering not only correctness but also precision.

\subsection{Accuracy}
The first metric we consider is \emph{Accuracy}, the percentage of responses that convey the correct chart-derived fact while permitting minor surface-form differences (e.g., casing, brief descriptors, small rounding). We apply light normalization (e.g., ignore surrounding punctuation/quotes, accept common month abbreviations), and map basic number words to digits: \texttt{zero} $\equiv$ \texttt{0}. The goal is to measure practical correctness rather than strict string identity.  A stricter exact match (EM) metric is reported separately.

\begin{table}[ht]
\centering
\caption{Accuracy (\%) for each prompting strategy and model (N=1{,}200 evaluation items)}
\label{tab:acc}
\begin{tabular}{|l|c|c|c|c|}
\hline
\textbf{Prompting Strategy} & \textbf{GPT-3.5} & \textbf{GPT-4} & \textbf{GPT-4o} & \textbf{Average} \\ \hline
ZSP     & 68.4 & 71.2 & 69.8 & 69.8 \\ \hline
FSP     & 72.1 & 74.5 & 70.3 & 72.3 \\ \hline
ZS-CoT  & 70.9 & 73.6 & 72.1 & 72.2 \\ \hline
FS-CoT  & 75.8 & 76.9 & 78.2 & 77.0 \\ \hline
\textbf{Average} & 71.8 & 74.1 & 72.6 &  \\ \hline

\end{tabular}
\end{table}

\subsection{Exact Match}
The second metric is \emph{Exact Match (EM)}. This measures the percentage of responses that match the expected answer string exactly, without any variation in words, symbols, or formatting. While similar to Accuracy, EM applies stricter criteria, making it especially useful for automated evaluation workflows that require consistent formatting.
\begin{table}[ht]
\centering

\caption{Exact Match (\%) for each prompting strategy and model (N=1{,}200 evaluation items)}
\label{tab:EM}
\begin{tabular}{|l|c|c|c|c|}
\hline
\textbf{Prompting Strategy} & \textbf{GPT-3.5} & \textbf{GPT-4} & \textbf{GPT-4o} & \textbf{Average} \\ \hline
ZSP & 60.9 & 62.1 & 59.1 & 60.7 \\ \hline
FSP & 65.2 & 60.4 & 68.5 & 64.7 \\ \hline
ZS-CoT & 62.3 & 61.2 & 60.7 & 61.4 \\ \hline
FS-CoT& 58.0 & 57.2& 58.4& 57.9\\ \hline
\textbf{Average} & 61.6 & 60.2 & 61.7 & \\ \hline
\end{tabular}
\end{table}

\subsection{Per-Category Analysis}\label{sec:per-category}
We also report Accuracy and EM by reasoning category (\emph{Arithmetic, Comparative, Boolean, and Direct Retrieval}). Values are macro-averaged across models.

\begin{table}[ht]
\centering
\caption{Accuracy and Exact Match (\%) by reasoning category (N = 300 per category, macro-averaged across models).}

\begin{tabular}{|l|c|c|c|c|}
\hline
\textbf{Prompting} & \textbf{Arithmetic} & \textbf{Comparative} & \textbf{Boolean} & \textbf{Direct Retrieval} \\ \hline
\multicolumn{5}{|c|}{\textbf{Accuracy}} \\ \hline
ZSP         & 62.3 & 70.8 & 72.4 & 75.6 \\ \hline
FSP         & 66.7 & 73.5 & 73.9 & 77.2 \\ \hline
ZS-CoT      & 64.1 & 72.6 & 74.2 & 76.8 \\ \hline
FS-CoT& 74.5 & 77.9 & 76.3 & 81.1 \\ \hline
\multicolumn{5}{|c|}{\textbf{Exact Match}} \\ \hline
ZSP         & 52.6 & 61.2 & 62.8 & 68.4 \\ \hline
FSP         & 58.3 & 65.7 & 65.1 & 71.9 \\ \hline
ZS-CoT      & 54.7 & 62.4 & 63.6 & 69.3 \\ \hline
FS-CoT& 53.2 & 56.8 & 55.4 & 62.7 \\ \hline
\end{tabular}
\end{table}

Together, these metrics provide a well-rounded view of system performance. Accuracy and Exact Match capture the core ability of a model to produce correct answers. By considering these perspectives together, we can identify not only which prompting strategy performs best overall, but also how different approaches trade off between strict correctness, informational richness, and presentation quality.

\section{Results and Discussion}

Our evaluation reveals clear trade-offs among prompting strategies for chart question answering. Few-Shot Chain-of-Thought (FS-CoT) achieved the highest average semantic accuracy (77.0\%), outperforming Few-Shot (72.3\%), Zero-Shot CoT (72.2\%), and Zero-Shot (69.8\%). This shows that reasoning traces encourage deeper analysis. However, FS-CoT also produced the lowest average exact match scores (57.9\%), indicating that while models often reason correctly, they struggle to return outputs in the precise expected format.  

In contrast, Few-Shot prompting gave the best average exact match (64.7\%) with competitive accuracy, showing exemplars help enforce formatting. Zero-Shot CoT offered modest gains over Zero-Shot at minimal cost, while plain Zero-Shot remained resource-efficient but less precise and consistent. 

FS-CoT responses were 2.5--3$\times$ longer than Zero-Shot, potentially limiting latency- or cost-sensitive use. Few-Shot increased cost by only $\sim$30\% while offering strong gains in both accuracy and EM, making it the most balanced strategy for real-world use. ZS-CoT provided modest improvements at negligible additional cost, reinforcing its role as a lightweight compromise. 

At the model level, GPT-4 achieved the highest average accuracy (74.1\%), though gains over GPT-3.5 (71.8\%) and GPT-4o (72.6\%) were modest. Notably, GPT-4o with FS-CoT reached 78.2\% accuracy, showing that smaller efficiency-oriented models can rival larger ones when guided by structured reasoning. The consistent 5--10 point gap between accuracy and exact match highlights a key challenge: models often capture the correct content but fail to output standardized formats, underscoring the importance of prompt design and post-processing for reliable deployment.

Beyond the quantitative results, our study shows that prompting strategies behave differently in chart reasoning compared to textual QA. Chart QA requires both symbolic computation and strict answer formatting, making it sensitive to reasoning depth and output consistency. Our error analysis reflects this: Zero-Shot often struggled with arithmetic aggregation, while FS-CoT produced answers that were logically correct but formatted incorrectly. These patterns suggest that chart reasoning reveals weaknesses in LLM prompting that do not appear in text-only tasks, underscoring the need for domain-specific evaluation.

At the time of experimentation, GPT-3.5-turbo, GPT-4, and GPT-4o supported 16k, 32k, and 128k-token context windows, respectively. While OpenAI has not disclosed parameter counts or exact latencies, GPT-4o is observed to be faster and more cost-efficient than GPT-4, with GPT-3.5 remaining the most lightweight. Latency varies with hardware, batching, and API configuration, providing practical guidance for model selection based on efficiency and accuracy trade-offs.

Our approach assumes access to structured chart inputs (tables/JSON) and bypasses challenges of parsing raw images, where OCR and grounding errors occur. It also relies on general-purpose LLMs without domain-specific fine-tuning, which may limit complex reasoning. Additionally, our evaluation was limited to tabular data and short questions.

\section{Conclusion and Future Work}
Our study demonstrates that prompting strategies play a crucial role in chart-based question answering. While FS-CoT achieves the highest semantic accuracy, its heavy token cost limits practicality in real-world deployments. Few-Shot prompting, by contrast, offers the best balance of accuracy and exact match consistency at a fraction of the cost, and ZS-CoT serves as a lightweight middle ground. These findings suggest that prompt design is not only a performance lever but also a cost–efficiency trade-off that practitioners must navigate. Beyond performance numbers, the persistent gap between semantic accuracy and exact match highlights an unmet need: models often reason correctly yet fail at strict output formatting. Addressing this challenge through prompt standardization, post-processing, or model-level calibration will be critical for reliable integration of LLMs into automated chart reasoning systems.

Future directions include building end-to-end pipelines for raw images, retrieval augmented and adaptive few-shot prompting, more diverse datasets, and lightweight fine-tuning (e.g., LoRA). We also plan to explore explainability and uncertainty estimation for greater transparency in high-stakes use cases.
%
%
%

\bibliographystyle{splncs04}
\typeout{} 
\bibliography{ref}

\end{document}